\documentclass{article}
\usepackage{ijcai24}

\usepackage{times}
\usepackage{soul}
\usepackage{url}
\usepackage[hidelinks]{hyperref}
\usepackage[utf8]{inputenc}
\usepackage[small]{caption}
\usepackage{graphicx}
\usepackage{amsmath}
\usepackage{amsthm}
\usepackage{booktabs}
\usepackage{algorithm}
\usepackage{algorithmic}
\usepackage[switch]{lineno}

\usepackage{arydshln} 
\usepackage[table]{xcolor}
\usepackage{colortbl}
\usepackage{graphicx}
\usepackage{amsmath}
\usepackage{amssymb}
\usepackage{booktabs}
\usepackage{enumitem}
\usepackage{xcolor}
\newcommand\crule[3][black]{\textcolor{#1}{\rule{#2}{#3}}}
\usepackage{booktabs, multirow} 
\definecolor{roadcolor}{RGB}{234,51,246}
\definecolor{sidewalkcolor}{RGB}{68,8,72}
\definecolor{parkingcolor}{RGB}{241,156,249}
\definecolor{othergroundcolor}{RGB}{160,32,76}
\definecolor{buildingcolor}{RGB}{246,202,69}
\definecolor{carcolor}{RGB}{111,149,238}
\definecolor{truckcolor}{RGB}{74,32,172}
\definecolor{bicyclecolor}{RGB}{136,227,242}
\definecolor{motorcyclecolor}{RGB}{37,59,146}
\definecolor{othervehiclecolor}{RGB}{96,81,242}
\definecolor{vegetationcolor}{RGB}{79, 173, 50}
\definecolor{trunkcolor}{RGB}{126, 65, 22}
\definecolor{terraincolor}{RGB}{171, 238, 105}
\definecolor{personcolor}{RGB}{234, 60, 49}
\definecolor{bicyclistcolor}{RGB}{234, 66, 195}
\definecolor{motorcyclistcolor}{RGB}{138, 42, 90}
\definecolor{fencecolor}{RGB}{238, 128, 69}
\definecolor{polecolor}{RGB}{252, 241, 161}
\definecolor{trafficsigncolor}{RGB}{233, 51, 35}
\definecolor{color1}{RGB}{176, 36, 24}
\definecolor{color2}{RGB}{119,185,0}
\definecolor{color3}{RGB}{0, 0, 200}
\definecolor{colorofteaser}{RGB}{176, 36, 24}

\definecolor{LightGrey}{rgb}{.9,.9,.9}
\definecolor{White}{rgb}{1.,0.,1.}
\definecolor{first}{rgb}{.8,.0,.0}
\definecolor{second}{rgb}{.0,.6,.0}
\definecolor{third}{rgb}{.0,.0,.8}
\definecolor{ceiling}{RGB}{214,  38, 40}   %
\definecolor{floor}{RGB}{43, 160, 4}     %
\definecolor{wall}{RGB}{158, 216, 229}  %
\definecolor{window}{RGB}{114, 158, 206}  %
\definecolor{chair}{RGB}{204, 204, 91}   %
\definecolor{bed}{RGB}{255, 186, 119}  %
\definecolor{sofa}{RGB}{147, 102, 188}  %
\definecolor{table}{RGB}{30, 119, 181}   %
\definecolor{tvs}{RGB}{160, 188, 33}   %
\definecolor{furniture}{RGB}{255, 127, 12}  %
\definecolor{objects}{RGB}{196, 175, 214} %
\definecolor{car}{rgb}{0.39215686, 0.58823529, 0.96078431}
\definecolor{bicycle}{rgb}{0.39215686, 0.90196078, 0.96078431}
\definecolor{motorcycle}{rgb}{0.11764706, 0.23529412, 0.58823529}
\definecolor{truck}{rgb}{0.31372549, 0.11764706, 0.70588235}
\definecolor{other-vehicle}{rgb}{0.39215686, 0.31372549, 0.98039216}
\definecolor{person}{rgb}{1.        , 0.11764706, 0.11764706}
\definecolor{bicyclist}{rgb}{1.        , 0.15686275, 0.78431373}
\definecolor{motorcyclist}{rgb}{0.58823529, 0.11764706, 0.35294118}
\definecolor{road}{rgb}{1.        , 0.        , 1.        }
\definecolor{parking}{rgb}{1.        , 0.58823529, 1.        }
\definecolor{sidewalk}{rgb}{0.29411765, 0.        , 0.29411765}
\definecolor{other-ground}{rgb}{0.68627451, 0.        , 0.29411765}
\definecolor{building}{rgb}{1.        , 0.78431373, 0.        }
\definecolor{fence}{rgb}{1.        , 0.47058824, 0.19607843}
\definecolor{vegetation}{rgb}{0.        , 0.68627451, 0.        }
\definecolor{trunk}{rgb}{0.52941176, 0.23529412, 0.        }
\definecolor{terrain}{rgb}{0.58823529, 0.94117647, 0.31372549}
\definecolor{pole}{rgb}{1.        , 0.94117647, 0.58823529}
\definecolor{traffic-sign}{rgb}{1.        , 0.        , 0.    }

\definecolor{barrier1}{RGB}{112,128,144}
\definecolor{bicycle1}{RGB}{220,20,60}
\definecolor{bus1}{RGB}{255, 127, 80}
\definecolor{car1}{RGB}{255, 158, 0}
\definecolor{const. veh.1}{RGB}{233, 150, 70}
\definecolor{motorcycle1}{RGB}{255,61,99}
\definecolor{pedestrian1}{RGB}{0,0,230}
\definecolor{traffic cone1}{RGB}{47,79,79}
\definecolor{trailer1}{RGB}{255,140,0}
\definecolor{truck1}{RGB}{255,99,71}
\definecolor{drive. suf.1}{RGB}{0,207,191}
\definecolor{other flat1}{RGB}{175,0,75}
\definecolor{sidewalk1}{RGB}{75,0,75}
\definecolor{terrain1}{RGB}{112,180,60}
\definecolor{manmade1}{RGB}{222,184,135}
\definecolor{vegetation1}{RGB}{0,175,0}

\title{Bridging Stereo Geometry and BEV Representation with\\ Reliable Mutual Interaction for Semantic Scene Completion}

\author{
Bohan Li$^{1,2}$
\and
Yasheng Sun$^3$\and
Zhujin Liang$^{4}$\and 
Dalong Du$^4$\and  \\
Zhuanghui Zhang$^4$ \and  
 Xiaofeng Wang$^5$ \and Yunnan Wang$^{1,2}$ \and Xin Jin$^{2,}$\thanks{Corresponding author.} \And Wenjun Zeng$^2$ \\
\affiliations
$^1$Shanghai Jiao Tong University, Shanghai, China.\\
$^2$Ningbo Institute of Digital Twin, Eastern Institute of Technology, Ningbo, China. \\
$^3$Tokyo Institute of Technology, Tokyo, Japan. 
$^4$PhiGent Robotics, Beijing, China. \\
$^5$Chinese Academy of Sciences, Beijing, China. \\
\emails
\{bohan\_li, wangyunnan\}@sjtu.edu.cn, sun.y.aj@m.titech.ac.jp,  \\
\{zhujin.liang, dalong.du, zhuanghui.zhang\}@phigent.ai, \\
wangxiaofeng2020@ia.ac.cn, \{jinxin, wenjunzengvp\}@eitech.edu.cn
}

\begin{document}

\twocolumn[{
\renewcommand\twocolumn[1][]{#1}%
\maketitle
\vspace{-0pt}
\begin{center}
	\begin{center}
		\includegraphics[width=15cm]{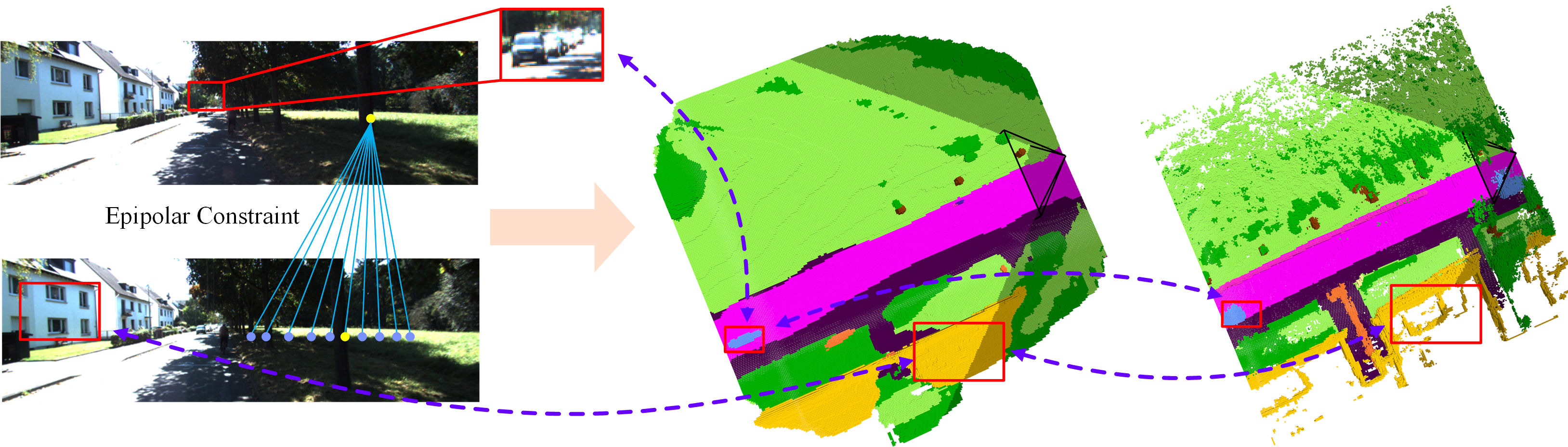}   
  
		\begin{tabular}{cccccc}	
			\multicolumn{6}{c}{
				\scriptsize
				\textcolor{bicycle}{$\blacksquare$}bicycle~
				\textcolor{car}{$\blacksquare$}car~
				\textcolor{motorcycle}{$\blacksquare$}motorcycle~
				\textcolor{truck}{$\blacksquare$}truck~
				\textcolor{other-vehicle}{$\blacksquare$}other vehicle~
				\textcolor{person}{$\blacksquare$}person~
				\textcolor{bicyclist}{$\blacksquare$}bicyclist~
				\textcolor{motorcyclist}{$\blacksquare$}motorcyclist~
				\textcolor{road}{$\blacksquare$}road~
				
}
    \\
    
			\multicolumn{6}{c}{
				\scriptsize
                    \textcolor{parking}{$\blacksquare$}parking~
				\textcolor{sidewalk}{$\blacksquare$}sidewalk~
				\textcolor{other-ground}{$\blacksquare$}other ground~
				\textcolor{building}{$\blacksquare$}building~
				\textcolor{fence}{$\blacksquare$}fence~
				\textcolor{vegetation}{$\blacksquare$}vegetation~
				\textcolor{trunk}{$\blacksquare$}trunk~
				\textcolor{terrain}{$\blacksquare$}terrain~
				\textcolor{pole}{$\blacksquare$}pole~
				\textcolor{traffic-sign}{$\blacksquare$}traffic sign			
			}
		\end{tabular}	
 	\end{center}
        \vspace{-0pt}
	\captionof{figure}{ \textbf{Overview of the proposed BRGScene}. The figure illustrates the stereo inputs, SSC prediction results and ground truth from left to right. We can see that our method shows promising performance in completing semantic scenes, especially for those challenging distant small objects, as indicated by the car highlighted with a red box.
        } 
\label{fig:teaser}
\end{center}
}]

\begin{abstract}
\vspace{-0pt}
3D semantic scene completion (SSC) is an ill-posed perception task that requires inferring a dense 3D scene from limited observations. Previous camera-based methods struggle to predict accurate semantic scenes due to inherent geometric ambiguity and incomplete observations. In this paper, we resort to \emph{stereo matching} technique and \emph{bird’s-eye-view (BEV)} representation learning to address such issues in SSC. Complementary to each other, stereo matching mitigates geometric ambiguity with epipolar constraint while BEV representation enhances the hallucination ability for invisible regions with global semantic context. However, due to the inherent representation gap between stereo geometry and BEV features, it is non-trivial to bridge them for \emph{dense prediction task} of SSC. Therefore, we further develop a unified occupancy-based framework dubbed \textbf{BRGScene}, which effectively \textbf{br}id\textbf{g}es these two representations with dense 3D volumes for reliable semantic \textbf{scene} completion. Specifically, we design a novel Mutual Interactive Ensemble (MIE) block for pixel-level reliable aggregation of stereo geometry and BEV features. Within the MIE block, a Bi-directional Reliable Interaction (BRI) module, enhanced with confidence re-weighting, is employed to encourage fine-grained interaction through mutual guidance. Besides, a Dual Volume Ensemble (DVE) module is introduced to facilitate complementary aggregation through channel-wise recalibration and multi-group voting.
Our method outperforms all published camera-based methods on SemanticKITTI for semantic scene completion. Our code is available on \url{https://github.com/Arlo0o/StereoScene}.
\vspace{-0pt}
\end{abstract}

\section{Introduction}
3D scene understanding is a fundamental task in computer vision~\cite{roberts1963machine}, facilitating a variety of applications such as autonomous driving, robotic navigation and augmented reality. Due to the limitations of real-world sensors such as restricted field of view, measurement noise, or sparse results, this task remains a challenging problem. To address this problem, 3D Semantic Scene Completion (SSC)~\cite{roldao20223d} is introduced to jointly predict the geometry and semantic segmentation of a scene. Given its inherent 3D nature, most existing SSC solutions~\cite{garbade2019two,roldao2020lmscnet,wu2020scfusion} employ 3D geometric signals, in the form of occupancy grids, point clouds, or distance fields, as their model inputs. Although they provide insightful geometric cues, it requires costly sensors (e.g. LiDAR) alongside considerable manual labor entailed in their deployment. Hence, it is worth exploring an efficient and effective approach for high-fidelity SSC solely with cost-friendly portable cameras. 

However, the absence of explicit 3D geometric information and incomplete observation pose large challenges to accurate geometry acquisition and reasonable hallucination in invisible regions
~\cite{cao2022monoscene,huang2023tri,li2023voxformer}. Thus, previous camera-based SSC solutions~\cite{cao2022monoscene,huang2023tri} tend to utilize learning-based projection techniques to convert 2D image features into a 3D dense space, but their predictions inevitably fall short of capturing accurate geometry without explicit constraints.
Later studies~\cite{li2023voxformer} attempt to introduce depth information to augment query for reliable geometry prediction. But their results still struggle to hallucinate reasonable invisible regions without ensembling global semantic context.

As a pivotal technique in 3D vision applications, stereo matching leverages explicit epipolar constraint to establish pixel-level correspondence, which is suitable for reconstructing dense 3D scene geometry~\cite{guo2019group}. On the other hand, the remarkable global robustness and hallucination ability of bird's-eye-view (BEV) representation, coupled with rich context and global semantic information have led to its widespread utilization in the 3D object detection community~\cite{philion2020lift,li2022bevformer,roddick2020predicting}. Inspired by the above two aspects, recent works have begun to simultaneously use stereo matching and BEV features for 3D perception. For instance, in the 3D object detection area, BEVStereo~\cite{li2022bevstereo} fuses monocular and temporal stereo depth maps before generating BEV features. BEVDepth~\cite{li2023bevdepth} improves geometry modeling with BVE representations by introducing extra depth supervision.
These strategies significantly enhance the effectiveness of sparse perception tasks (e.g. 3D object detection) by focusing on coarse-grained predictions at the region level for common visual objects.\looseness=-1

Despite remarkable advances in 3D object detection, it's non-trivial in the semantic scene completion (SSC) task to bridge the representation gap between stereo geometry and BEV features within a unified framework for pixel-level reliable prediction. This difficulty arises due to the structural variations among similar objects and the insufficient region-level coarse information for pixel-level semantics and geometry in complex real-world scenarios.

Given these concerns, we propose \textbf{BRGScene}, a framework that bridges stereo matching technique and BEV representation for fine-grained reliable SSC, and the results are shown in Figure~\ref{fig:teaser}.
Our framework aims to fully exploit the potential of vision inputs with explicit geometric constraint of stereo matching~\cite{guo2019group,jiankun2022crestereo} and global semantic context of BEV representation~\cite{philion2020lift,li2023bevdepth}.
Different from previous methods that focus on 2D features or depth maps~\cite{li2023voxformer,li2023bevdepth}, we propose to employ the {dense 3D volumes} of stereo and BEV representations for SSC.

Given the distinct nature of the two volumes, we devise a \emph{Mutual Interactive Ensemble (MIE)} block to bridge the gap for fine-grained reliable perception. Specifically, a \emph{Bi-directional Reliable Interaction (BRI)} module is designed to guide each volume to retrieve pixel-level reliable information. A confidence re-weighting strategy inspired by MVS~\cite{chen2020mvsnet++} is incorporated on top of the BRI module to further enhance the performance. Furthermore, a \emph{Dual Volume Ensemble (DVE)} module is introduced to facilitate complementary aggregation with channel-wise recalibration and multi-group feature voting.
Our contributions are summarized as follows: \textbf{1)} We propose a novel framework that resorts to both stereo and BEV representations with dense 3D volumes for precise geometry modeling and hallucination ability enhancement in SSC. \textbf{2)} To bridge the representation gap for fine-grained reliable perception, a \emph{Mutual Interactive Ensemble} block is designed to take advantage of the complementary merits of the volumes in the two representations. \textbf{3)} Our proposed BRGScene outperforms state-of-the-art VoxFormer-T with a 14.5\% relative improvement on the SemanticKITTI leaderboard.\looseness=-1

\section{Related Works}

\subsection{Semantic Scene Completion}

Semantic scene completion is a dense 3D perception task that jointly estimates semantic segmentation and scene completion~\cite{behley2019semantickitti,cai2021semantic}. 
To provide additional texture or geometry information, some works~\cite{cai2021semantic,li2019rgbd} exploit multi-modal inputs, such as RGB images coupled with geometric cues. 
Another slew of studies~\cite{cao2022monoscene,li2023voxformer} aims to achieve semantic scene completion solely with camera-only inputs. 
For instance, MonoScene~\cite{cao2022monoscene} lifts a monocular image using 2D-3D projections and leverages 2D and 3D UNets for semantic scene completion.
TPVFormer~\cite{huang2023tri} utilizes a tri-perspective view representation and attention mechanism for 3D scene understanding.
VoxFormer~\cite{li2023voxformer} employs a transformer-based framework where a sparse set of depth-based voxel queries are devised for scene structure reconstruction.\looseness=-1

\subsection{Stereo Matching Based 3D Perception}
With the advances of deep convolution neural networks, the quality of depth predictions from stereo images~\cite{poggi2022stereodepth} has steadily improved and led to a remarkable improvement in downstream 3D vision applications such as object detection, surface reconstruction and augmented reality. 
GC-Net \cite{kendall2017end} first proposes to employ 3D CNNs in the stereo matching framework, where 2D corresponding features are mapped to a 3D cost volume through a concatenation operation. 
RAFTStereo~\cite{Lahav2021raftstereo} and CREStereo\cite{jiankun2022crestereo} utilize feature correlation to produce matching cost volume, which is subsequently optimized through sequential refinement modules for depth prediction. 
However, in conditions such as occlusion and large textureless regions, stereo matching prediction performance drops significantly.

\begin{figure*}[!ht]
\vspace{-0pt}
\hsize=\textwidth %
\centering
\includegraphics[width=1\textwidth]{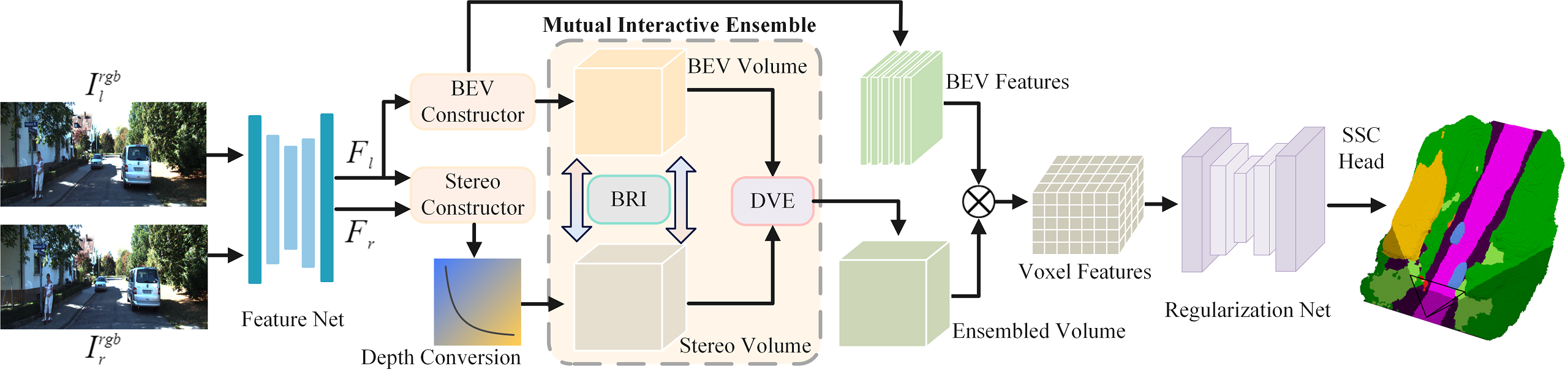}
\caption{ \textbf{Overall framework of our proposed BRGScene}. Given input stereo images, we employ 2D UNet to extract image features. The BEV latent volume and stereo geometric volume are constructed by a \emph{BEV Constructor} and a \emph{Stereo Constructor}, respectively. To bridge their representation gap for fine-grained
reliable perception, a \emph{Mutual Interactive Ensemble} block is proposed to take advantage of complementary merits of the volumes.}
\label{figoverall}
 \vspace{-0pt}
\end{figure*}

\subsection{Bird's-Eye-View Representation }
The bird's-eye-view is a widely used representation in 3D object detection since it provides a clear depiction of the layout and strong hallucination ability from a top-down perspective~\cite{philion2020lift,hu2021fiery}. 
Lift-Splat~\cite{philion2020lift} extracts BEV representations from an arbitrary number of cameras by implicitly unprojecting 2D visual inputs based on estimated depth distribution. 
BEVDepth~\cite{li2023bevdepth} leverages a camera-aware monocular depth estimation module to improve depth perception in BEV-based 3D detection.  
However, region-level coarse perception in the object detection task is not effective enough for the dense prediction task of SSC.
In this work, our objective is to investigate reliable pixel-level prediction for high-quality semantic scene completion.

\section{Methodology}
We present our proposed BRGScene that aims to jointly infer dense 3D geometry and semantics solely from RGB images. In this section, we first introduce a hybrid occupancy-based SSC framework (Sec.~\ref{sec:preliminary}), including problem formulation and architecture overview. Then we provide detailed construction of \emph{dual volume} representations (Sec.~\ref{sec32}). To bridge their representation gap for fine-grained
reliable perception, we depict our devised ensemble block (Sec.~\ref{sec:integration}). Finally, we introduce our SSC generator and training paradigm (Sec.~\ref{sec:ssc}). \looseness=-1

\subsection{The Proposed Framework of BRGScene}

We propose BRGScene, a framework that bridges the stereo matching technique and the BEV representation for fine-grained reliable SSC.

\label{sec:preliminary}
\noindent \textbf{Problem Formulation.} Given a set of stereo RGB images $I^{rgb}_{Stereo}=\{I^{rgb}_l, I^{rgb}_r\}$, our goal is to jointly infer geometry and semantics of a 3D scene. The scene is represented as a voxel grid $\textbf{Y} \in \mathbb{R}^{H\times W \times Z}$, where $H,W,Z$ denote the height, width and depth in 3D space. Regarding each voxel, it will be assigned to a unique semantic label belonging to $C \in \{ c_0, c_1, \cdots, c_M \}$, which either occupies empty space $c_0$ or falls on a specific semantic class $\{c_1, c_2, \cdots, c_M \}$. Here $M$ denotes the total number of semantic classes. We would like to learn a transformation $\hat{\textbf{Y}}=\Theta(I^{rgb}_{Stereo})$ to approach ground truth 3D semantics $\textbf{Y}$.

\noindent \textbf{Architecture Overview}. The overall architecture of our proposed framework is illustrated in Figure~\ref{figoverall}. We follow a common paradigm~\cite{cao2022monoscene} that employs successive 2D and 3D UNets as backbones. The input stereo images $I^{rgb}_{Stereo}$ are separately encoded by a 2D UNet into paired context-aware features ${\textbf{F}_l}$ and ${\textbf{F}_r}$. Then we leverage a \emph{Stereo Constructor} to convert these features into a dense 3D volume $\textbf{V}_{Stereo} \in \mathbb{R}^{D \times H \times W}$. In parallel, a \emph{BEV Constructor} lift 2D features ${\textbf{F}_l}$ of the left image to a latent BEV volume $\textbf{V}_{BEV} \in \mathbb{R}^{D \times H \times W}$ alongside its context feature $\textbf{C}_{BEV} \in \mathbb{R}^{C_b \times H \times W}$ following standard protocol of ~\cite{philion2020lift}. Sequentially, the two-stream built volumes are bridged and aggregated to a new volume $\textbf{V}_{ens}$ by a \emph{Mutual Interactive Ensemble} block. Finally, the context features $\textbf{C}_{BEV}$ splat along volume $\textbf{V}_{ens}$ by outer-product, which will be fed to a 3D UNet for semantic segmentation and completion. \looseness=-1

\subsection{Dual Volume Construction} 
\label{sec32}

Unlike previous studies~\cite{li2023voxformer,li2023bevdepth} which focus on 2D representations, we employ \emph{3D volumetric representation} to resolve dense scene understanding. Specially, we introduce hybrid volumetric representations with stereo and BEV volumes to take full advantage of camera inputs. \looseness=-1

\noindent \textbf{2D Feature Extraction Backbone}. For image feature extraction, 2D UNet with pre-trained EfficientNetB7~\cite{tan2019efficientnet} is leveraged to separately process left and right input images. Note that we utilize shared weights to encourage efficient correspondence feature learning.

\noindent \textbf{Stereo Geometric Volume Constructor}. With the obtained unary features ${\textbf{F}_l}$ and ${\textbf{F}_r}$ from the left and right images, \emph{Stereo Constructor} targets to build a stereo depth volume $\textbf{V}_{Stereo}$ by matching them with epipolar constraint. Specifically, group-wise correlation~\cite{guo2019group} is first adopted to generate disparity cost volume. Formally,
\begin{equation}\label{a}
D_{gwc}(d,x,y,g)=\frac{1}{N_{c}/N_{g}} \left< f_{g}^{l}(x,y), f_{g}^{r}(x-d,y)\right>,
\end{equation}
where $ \left< \cdot,\cdot\right>$represents the inner product, $N_{c}$ is the channels of input features, $N_{g}$ is the number of groups, $f_{g}^{l}$ and $f_{g}^{r}$ represent $g^{th}$ left and right feature group, respectively. 
Afterward, the disparity volume is converted into a depth volume following~\cite{you2019pseudo}, which is formulated as:
\begin{equation} \label{b}
z_{(u,v)}=\frac{f_{u}\times b}{D_{(u,v)}}, x=\frac{(u-c_{u})\times z}{f_{u}}, y=\frac{(v-c_{v})\times z}{f_{v}},
\end{equation}
where $f_{u}$ and $f_{v}$ represent the horizontal and vertical focal length, ($c_{u}, c_{v}$) is the camera center. 
Next, we employ 3D CNNs following~\cite{guo2019group} for dimension reduction and finally squeeze the channel dimension to construct the dense stereo depth volume $\textbf{V}_{Stereo} \in \mathbb{R}^{ D \times H \times W} $.

\noindent \textbf{BEV Latent Volume Constructor}. Although the stereo constructor provides accurate estimation in matched regions, it struggles in extreme conditions where severe occlusion or high reflection happens. Unlike the stereo-based approach relying on strict geometric matching, BEV representations
are obtained by lifting an image $I^{rgb}$ to a shared bird's eye space through 3D prior. Following \cite{philion2020lift,li2023bevdepth}, we feed visual features $\textbf{F}_l$ to a neural network and obtain a latent depth distribution $\textbf{V}_{BEV} \in \mathbb{R}^{D \times H \times W}$ with its associated context features $\textbf{C}_{BEV} \in \mathbb{R}^{C_b \times H \times W}$. Since this distribution is essentially a voxel grid that stores the probability of all possible depths, we denote it as BEV latent volume for the sake of clarity.

\subsection{Mutual Interactive Ensemble}\label{sec34}
\label{sec:integration}
To achieve fine-grained reliable perception, the Mutual Interactive Ensemble (MIE) block is introduced to bridge the representation gap between stereo volume $\textbf{V}_{Stereo}$ and BEV volume $\textbf{V}_{BEV}$ by mutually reinforcing each other and integrating their respective potentials at pixel level.

\begin{figure}[!ht]
\vspace{0pt}
	\begin{center}
		\includegraphics[width=0.8\linewidth]{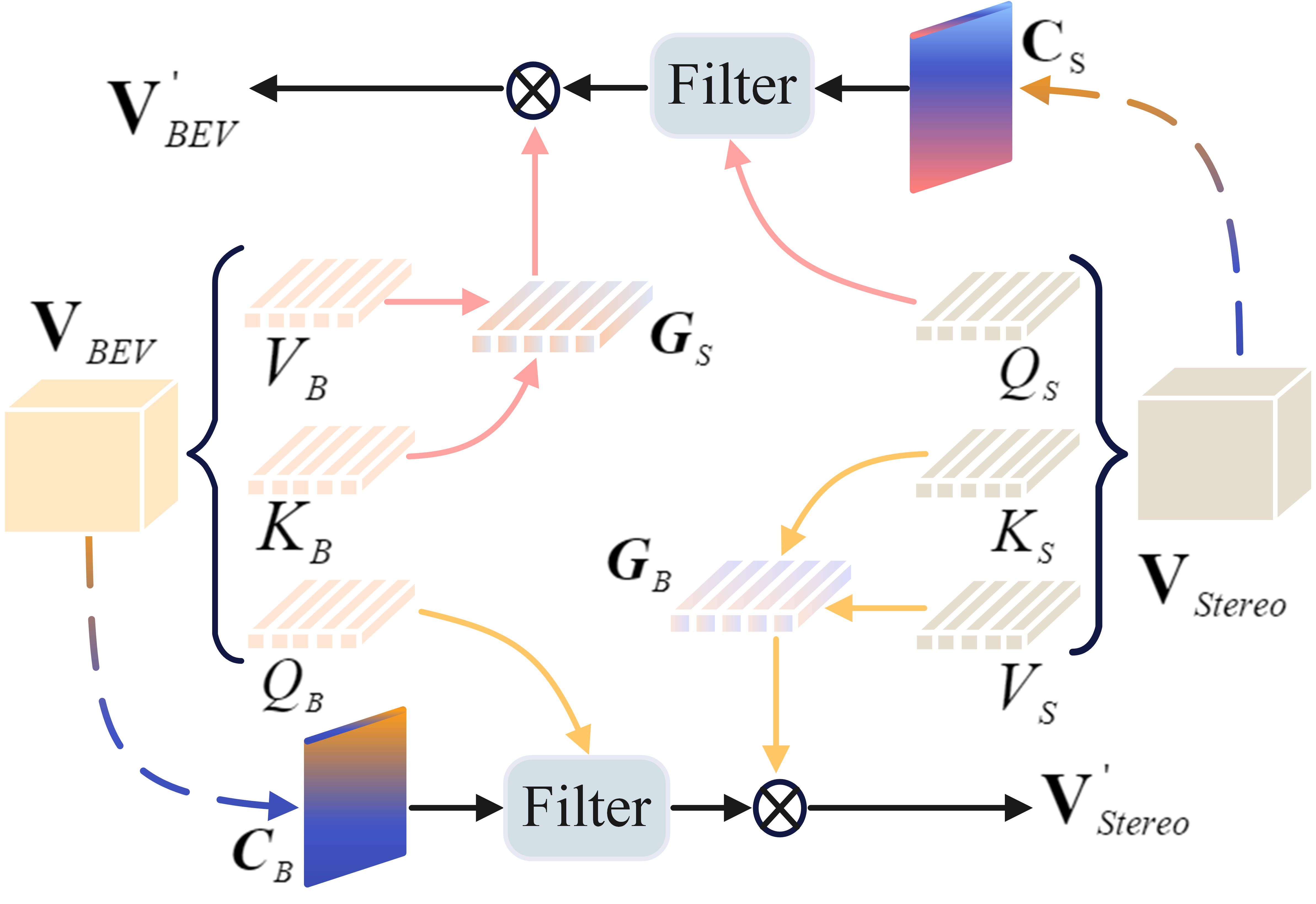}  
 	\end{center}
        \vspace{-0pt}
	\caption{ The structure of the proposed Bi-directional Reliable Interaction module, which is designed for pixel-level reliable geometry information interaction.} 
	\label{figblock}
 \vspace{-0pt}
\end{figure}

\subsubsection{Bi-directional Reliable Interaction.}
For pixel-level reliable interaction, we propose an initial interactive stage that selectively retrieves dependable information alongside its counterpart volume. More specifically, a Bi-directional Reliable Interaction (BRI) module as shown in Figure~\ref{figblock} is devised to interactively guide reliable predictions of its contrary side through a cross-attention mechanism. For stereo volume $\textbf{V}_{Stereo}$, we first obtain its query $Q_S$, key $K_S$ and value $V_S$ by flattening in spatial and depth dimensions following standard protocol~\cite{wang2018non,liao2022wt}.
Similarly, the BEV volume $\textbf{V}_{BEV}$ is forwarded and its query, key and value are denoted as $Q_B, K_B, V_B$, respectively.\looseness=-1 

Then we construct the interacted volume with the cross-attention operation. To reduce computational and memory consumption, we follow~\cite{katharopoulos2020transformers,kitaev2020reformer} to compute linear cross-attention.
Specially, the interacted BEV volume $\textbf{V}_{BEV}'$ is obtained by:

\begin{equation} \label{eqca}
\begin{split}
\textbf{V}_{BEV}' &= CrossAtt(Q_{S},K_{B},V_{B} )  \\
                    & = \phi_q(Q_{S}) \textbf{G}_{B} = \phi_q(Q_{S}) (\phi_k{ (K_{B}) }^{T} V_{B} ),
\end{split}
\end{equation}

where $\phi_q$ and $\phi_k$ denote the softmax function along each row and column of the input matrix, respectively. $\textbf{G}_{B}$ represents global contextual vectors of the BEV representation.
In this way, $\textbf{V}_{BEV}'$ retrieves the relevant aspects of the stereo sides, thereby providing an alternative perspective on the feature importance of the BEV side. 
Likewise, its opposite interacted volume $\textbf{V}_{Stereo}'$ is computed by $CrossAtt(Q_{B},K_{S},V_{S} )$ to encourage reliable geometry information exchange.

\emph{Depth Confidence Filtering}. In order to further retrieve pixel-level reliable information for dense prediction, we develop a depth confidence filtering strategy, which explicitly takes advantage of the involved reliable geometry information behind the volume. We aim to utilize its depth confidence information to enforce the cross-attention operation similar to \cite{chen2020mvsnet++}. Particularly,  
to project the volume to a confidence map $\textbf{C}_S$ 
, we first adopt $softmax$ to convert depth cost value $d_i$ into a probability form, and then take out the highest probability value among all depth hypothesis planes along the depth dimension as the prediction confidence. The process is formally written as:
\begin{equation}
{\textbf{C}_S= WTA(\phi ( \textbf{V}_{Stereo} )) = WTA \left\{   \frac{\exp(d_i)} 
 {\sum_{j=1}^{D_{max}}\exp(d_j)}  \right\},}
\end{equation}
where the $softmax$ is applied across the depth dimension and $WTA$ represents winner-takes-all operation. $D_{max}$ denotes the length of the depth dimension. 

Next, we revisit the cross-attention operation in Equation~\ref{eqca} and construct pixel-level reliable retrieval with the confidence map $\textbf{C}_S$ to identify the criteria for an optimized formulation:

\begin{equation} \label{}
{ CrossAtt(Q_{S},K_{B},V_{B} ) =  \phi_q(Q_{S}) \odot \textbf{C}_S ( \phi_k{ (K_{B}) }^{T} V_{B} ),}
\end{equation}

where $\odot$ represents the element-wise product, through which the reliable geometry information is preserved while low-confidence information is suppressed.

\begin{figure}[!ht]
\vspace{0pt}
\includegraphics[width=1\linewidth]{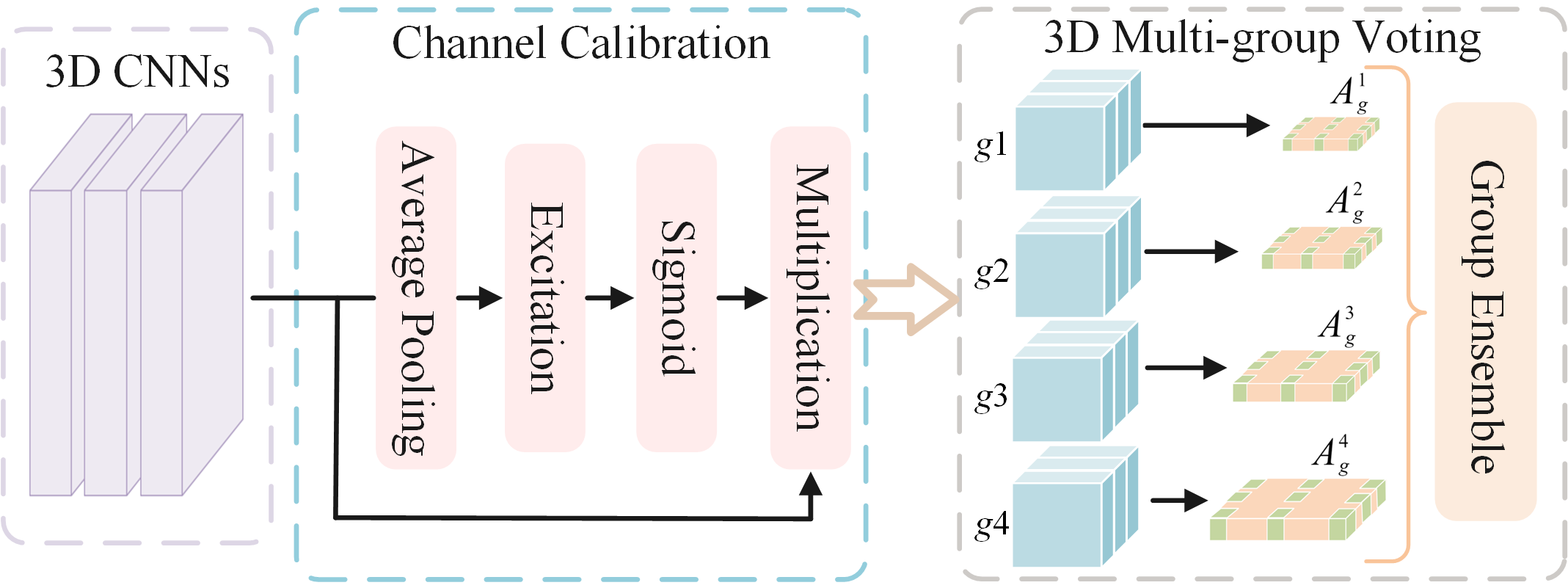}  
\caption{The structure of the proposed Dual Volume Ensemble module, which is devised for mutually beneficial aggregation.} 

\label{figblock2}
\vspace{-0pt}
\end{figure} 

\subsubsection{Dual Volume Ensemble.}
With the interacted volume representations $\textbf{V}_{Stereo}'\in\mathbb{R}^{1 \times D \times H \times W}$ and $\textbf{V}_{BEV}'\in\mathbb{R}^{1 \times D \times H \times W}$, the primary objective of this module is to leverage their strengths and facilitate mutually beneficial complementation. 

As illustrated in Figure~\ref{figblock2},
the DVE takes as input concatenated features $\textbf{V}'_{cat}=[\textbf{V}_{Stereo}', \textbf{V}_{BEV}']\in \mathbb{R}^{2 \times D \times H \times W}$ and outputs ensembled volume $\textbf{V}_{ens}$. 
Especially, the input $\textbf{V}'_{cat}$ is first fed into residual 3D CNNs for regularization and channel adjustment, which generates transformed representation $\textbf{V}_f \in \mathbb{R}^{C_f \times D \times H \times W}$. The transformed representation $\textbf{V}_f$ is further processed by Channel-wise Recalibration and Multi-group Voting, which are described in detail below.

 \emph{Channel-wise Recalibration.}
To fully exploit its contextual information~\cite{hu2018squeeze}, we utilize average pooling to squeeze the information into a channel descriptor $\textbf{z}_c$. More specifically, we shrink $\textbf{V}_f$ along both the depth dimension $D$ and spatial dimension $ H \times W$:
\begin{equation}
    \textbf{z}_c = \frac{1}{D\times H\times W} \sum_{d=1}^D \sum_{i=1,j=1}^{H,W} \textbf{V}_f(d,i,j),
\end{equation}

Subsequently, an excitation block~\cite{hu2018squeeze} is leveraged to capture its channel-wise dependencies. Formally, the channel descriptor $\textbf{z}_c$ is updated by two stacked bottleneck-shape convolutions with non-linear activation. Finally, the updated channel descriptor is employed to re-weight the previous transformed feature $\textbf{V}_f$ along the channel dimension: 
\begin{equation}
     \textbf{V}_f ' = \sigma(\textbf{W}_2\delta(\textbf{W}_1 \textbf{z}_c))\cdot  \textbf{V}_f ,
\end{equation}
where the ${W}_1$ and ${W}_2$ represent $ 1\times1\times1 $ convolutions with dimensionality-reduction. The $\delta$ denotes standard GELU activation and the $\sigma$ indicates the sigmoid gate.

 \emph{Multi-group Voting.}
 To further capture multi-scale context, we split $\textbf{V}_f'$ into four groups and employ 3D atrous convolutions~\cite{li2019rgbd,cao2022monoscene} with different dilation rates. 
The contextual information in different groups exhibits distinct receptive fields, staying non-interfering with each other.
Finally, we ensemble the multi-scale context between different voting groups to construct $\textbf{V}_{ens}$:\looseness=-1
\begin{equation}
\textbf{V}_{ens} = \mathbb{P} \left\{ \mathrm{Concat} \left( A_{g}^{i}(\mathrm{Split}_{channel}^{g1\sim g4}( \textbf{V}_f' )) \right) \right\},
\end{equation}
where $A_{g}^{i}$ denotes 3D atrous convolution employed in the $i^{th}$ context group. 
$\mathbb{P}$ is composed of point-wise convolution with GELU activation and group normalization. The point-wise convolution encourages channel-aware mixing by $1\times1\times1$ kernel size, through which the ensembled volume $\textbf{V}_{ens}$ takes into consideration voting features of different aspects.

\subsection{Semantic Scene Completion}
\label{sec:ssc}
To make use of this high-quality volume $\textbf{V}_{ens}$ for semantic scene completion, we augment it with its associated context information $\textbf{C}_{BEV}$. The extracted context information from input images is placed in a specific location of the bird-eye's representation by an outer product operation similar to~\cite{philion2020lift,li2023bevdepth}. Formally, the ensembled voxel features $\textbf{F}_{vox} \in \mathbb{R}^{C_b \times D \times H \times W}$ is computed by:
\begin{equation}
\textbf{F}_{vox} = \textbf{C}_{BEV} \otimes \textbf{V}_{ens},
\end{equation}
In this way, we are able to seamlessly blend the complementary benefits of stereo representation of precise geometry and BEV features of rich semantic context.

\noindent \textbf{Semantic Segmentation Learning.} Following~\cite{cao2022monoscene}, we leverage 3D UNet to regularize the ensembled voxel features. Its output features are fed to a SSC head holding upsampling and a softmax layer for semantic occupancy prediction $\hat{\textbf{Y}}$.

\noindent \textbf{Network Training.} 
We follow the basic learning objective of MonoScene~\cite{cao2022monoscene} for semantic scene completion. 
Standard semantic loss $\mathcal{L}_{\text{sem}}$ and geometry loss $\mathcal{L}_{\text{geo}}$ are leveraged for semantic and geometry supervision, while an extra class weighting loss $\mathcal{L}_{ce}$ is also added.
To further enforce the ensembled volume, we adopt a binary cross entropy loss $\mathcal{L}_{depth}$ to encourage the sparse depth distribution. The overall learning objective of this framework is formulated as:
\begin{equation}
   { \mathcal{L} = \mathcal{L}_{depth} + \mathcal{\lambda}_{ce} \mathcal{L}_{ce} + \mathcal{\lambda}_{sem} \mathcal{L}_{\text{sem}}+ \mathcal{\lambda}_{geo} \mathcal{L}_{\text{geo}} . }
\end{equation}
where several $\lambda$s are balancing coefficients.

\begin{table*}[!ht]
\vspace{-0pt}
\begin{center}
\small
\renewcommand\tabcolsep{5.2pt}
\begin{tabular}{l|ccccccc}
\toprule
\textbf{Methods}   & BRGScene(ours)  & VoxFormer-T & VoxFormer-S  & OccFormer & SurroundOcc & TPVFormer & MonoScene  \\ \midrule
\textbf{Input}  &  Stereo &  Stereo-T & Stereo &  Mono &  Mono  &  Mono &  Mono   \\ \midrule

\textbf{IoU (\%)} $\uparrow$      & \textbf{43.34}  & {43.21} & {42.95}  & 34.53  & 34.72 &34.25 &34.16    \\ \midrule
\textbf{mIoU (\%)} $\uparrow$     & \textbf{15.36}    & {13.41}  & 12.20 &12.32 & 11.86 &11.26   &11.08     \\ \midrule
\crule[carcolor]{0.13cm}{0.13cm} \textbf{car} (3.92\%)     & \textbf{22.80} & {21.70} & 20.80  & 21.60 &20.60 & 19.20   &  18.80  \\

\crule[bicyclecolor]{0.13cm}{0.13cm} \textbf{bicycle} (0.03\%)    & \textbf{3.40}  &{1.90} & 1.00& 1.50 &1.60 & 1.00 & {0.50}  \\

\crule[motorcyclecolor]{0.13cm}{0.13cm} \textbf{motorcycle} (0.03\%)   & \textbf{2.40} &{1.60} &0.70 & {1.70} &1.20 & 0.50  &  {0.70}    \\

\crule[truckcolor]{0.13cm}{0.13cm} \textbf{truck} (0.16\%)        & {2.80} & 3.60 &3.50 & {1.20} &1.40  & \textbf{3.70}  &  {3.30}  \\

\crule[othervehiclecolor]{0.13cm}{0.13cm} \textbf{other-veh.} (0.20\%)   & \textbf{6.10}  &4.10 & 3.70& 3.20 &{4.40} & 2.30  &{4.40} \\

\crule[personcolor]{0.13cm}{0.13cm} \textbf{person} (0.07\%)       & \textbf{2.90} &{1.60} &1.40 & {2.20} &1.40 &{1.10}  & {1.00}  \\

\crule[bicyclistcolor]{0.13cm}{0.13cm} \textbf{bicyclist} (0.07\%)    & {2.20} & 1.10 & \textbf{2.60} &1.10 &2.00  & {2.40} & {1.40}    \\

\crule[motorcyclistcolor]{0.13cm}{0.13cm} \textbf{motorcyclist} (0.05\%) & \textbf{0.50} & 0.00 &0.20 & 0.20 &0.10 & 0.30   &  {0.40}  \\

\crule[roadcolor]{0.13cm}{0.13cm} \textbf{road} (15.30\%)        & \textbf{61.90} &54.10 &53.90 & 55.90 &{56.90} &{55.10}    &{54.70}  \\

\crule[parkingcolor]{0.13cm}{0.13cm} \textbf{parking} (1.12\%)      & \textbf{30.70}  & 25.10& 21.10 &{31.50} &30.20 & {27.40}   & {24.80}   \\

\crule[sidewalkcolor]{0.13cm}{0.13cm} \textbf{sidewalk} (11.13\%)    & \textbf{31.20}  &26.90 &25.30 & {30.30} &28.30 & {27.20} & {27.10}  \\

\crule[othergroundcolor]{0.13cm}{0.13cm} \textbf{other-grnd} (0.56\%)    & \textbf{10.70} & {7.30} &5.60 &6.50 &6.80 & 6.50 & {5.70}  \\

\crule[buildingcolor]{0.13cm}{0.13cm} \textbf{building} (14.10\%)    & \textbf{24.20} &{23.50} & 19.80 &15.70 &15.20 & 14.80 & {14.40}     \\

\crule[fencecolor]{0.13cm}{0.13cm} \textbf{fence} (3.90\%)        & \textbf{16.50}  &{13.10} & 11.10 & 11.90 & 11.30 & {11.00} &  {11.10} \\

\crule[vegetationcolor]{0.13cm}{0.13cm} \textbf{vegetation} (39.3\%)   & {23.80}  & \textbf{24.40} & 22.40& 16.80 &14.90 & 13.90 & {14.90} \\

\crule[trunkcolor]{0.13cm}{0.13cm} \textbf{trunk} (0.51\%)        & \textbf{8.40} & {8.10} & 7.50 & 3.90 & 3.40 & 2.60   &2.40 \\

\crule[terraincolor]{0.13cm}{0.13cm} \textbf{terrain} (9.17\%)   & \textbf{27.00} &{24.20} &21.30 & 21.30 &19.30  &20.40 & {19.50}   \\

\crule[polecolor]{0.13cm}{0.13cm} \textbf{pole} (0.29\%)         & \textbf{7.00}  & {6.60} & 5.10 & 3.80 &3.90 & 2.90 &3.30  \\

\crule[trafficsigncolor]{0.13cm}{0.13cm} \textbf{traf.-sign} (0.08\%)   & \textbf{7.20} &{5.70} & 4.90 & 3.70 &2.40 & 1.50 &2.10  \\  \bottomrule

\end{tabular}
\vspace{-0pt}
\caption{\textbf{Quantitative results} on the SemanticKITTI test set. The top two performers are marked \textbf{bold}. Our method outperforms temporal stereo-based (Stereo-T) VoxFormer-T in terms of mIoU.\looseness=-1 }
\label{tabq1}        
\vspace{-0pt}
\end{center}
\end{table*}

\begin{table}[!ht]
\begin{center}
\small
\renewcommand\tabcolsep{3.0pt}
\begin{tabular}{l|ccc}
\toprule
 \textbf{Methods}& \textbf{Input}   & \textbf{mIoU (\%)} $\uparrow$ & \textbf{Time (s)} $\downarrow$      \\  \midrule
 SSCNet${}^{*}$ (2017) &  Stereo-PTS & 10.31 &\textbf{0.187}\\
 LMSCNet${}^{*}$ (2020) &  Stereo-PTS & 10.45 & 0.214 \\
MonoScene${}^{*}$  (2022)  & Stereo    &    12.82  &   {0.274}            \\
TPVFormer${}^{*}$  (2023)  &  Stereo   & 13.06 & 0.313   \\
OccFormer${}^{*}$  (2023)  & Stereo  &  13.57 & 0.338   \\

VoxFormer-S (2023)    &  Stereo   &   {12.35} &  {0.256}    \\ 
VoxFormer-T (2023)  &  Stereo-T  &   { 13.35 }  & 0.307   \\ 
\rowcolor{gray!10} BRGScene (ours) & Stereo & \textbf{15.43}  & 0.285 \\  
 \bottomrule
\end{tabular}
\vspace{-0pt}
\caption{ \textbf{Evaluation results of stereo variants} on the SemanticKITTI validation set. For MonoScene${}^{*}$ and TPVFormer${}^{*}$, We employ stereo images as inputs. For SSCNet${}^{*}$ and LMSCNet${}^{*}$, we leverage stereo depth net to generate pseudo point clouds (Stereo-PTS). }
\vspace{-0pt}
\label{tabq3}
\end{center}
\end{table}

\section{Experiments}
\subsection{Experimental Settings}
\noindent \textbf{Datasets.}
We evaluate the proposed \textbf{BRGScene} on SemanticKITTI~\cite{behley2019semantickitti} that is popularly used in a great number of previous studies. There are 22 driving outdoor scenes from the KITTI Odometry Benchmark~\cite{geiger2012we}, covering diverse and challenging autonomous driving situations. SemanticKITTI holds semantic annotations of LiDAR sweeps that are registered, aggregated and voxelized as 256$\times$256$\times$32 grid of 0.2m voxels. The target ground truth of each voxel grid is annotated as one of 21 classes (1 unknown, 1 free and 19 semantics). The SemanticKITTI benchmark provides both voxelized LiDAR scans and RGB images as model input options. We solely utilize RGB images since our main focus is to explore the portable camera-only signals as did in MonoScene~\cite{cao2022monoscene}. 

\noindent \textbf{Implementation Details.} We set the 3D UNet input to 128$\times$128$\times$16 (1:2) for efficient memory usage, whose feature will be upscaled to 1:1 at completion head by deconvolution operation. The $\lambda$s are empirically set to 1. Our model is implemented on PyTorch. The model is trained for 30 epochs using the AdamW optimizer~\cite{loshchilov2017decoupled} with a learning rate of $1\times10^{-4}$ and batch size set to 8.\looseness=-1

\noindent \textbf{Evaluation Metrics}.
Regarding quantitative evaluations, we conduct experiments on the typical metrics~\cite{li2023voxformer,cao2022monoscene} that have been widely employed in the field of SSC. Specially, we leverage \textbf{IoU} (Intersection over Union) to account for the scene completion (SC) task and \textbf{mIoU} (mean Intersection over Union) to measure the performance of the semantic scene completion (SSC) task, respectively. For both of these two metrics, higher values are desirable, where high IoU indicates accurate geometric prediction and high mIoU implies precise semantic segmentation.\looseness=-1

\begin{figure*}[!ht]
   \vspace{-0pt}
	\begin{center}
		\includegraphics[width=1\linewidth]{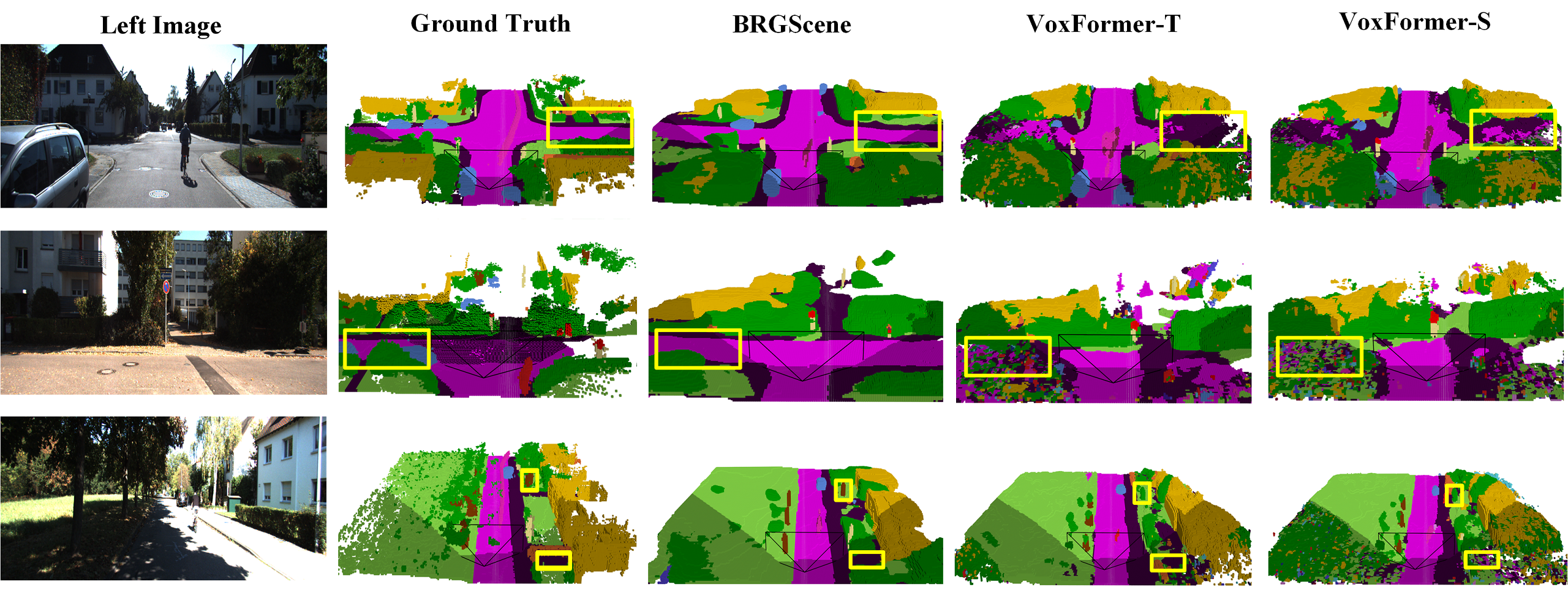}   
        \vspace{0pt}
		\begin{tabular}{cccccc}
			\multicolumn{6}{c}{
				\scriptsize
				\textcolor{bicycle}{$\blacksquare$}bicycle~
				\textcolor{car}{$\blacksquare$}car~
				\textcolor{motorcycle}{$\blacksquare$}motorcycle~
				\textcolor{truck}{$\blacksquare$}truck~
				\textcolor{other-vehicle}{$\blacksquare$}other vehicle~
				\textcolor{person}{$\blacksquare$}person~
				\textcolor{bicyclist}{$\blacksquare$}bicyclist~
				\textcolor{motorcyclist}{$\blacksquare$}motorcyclist~
				\textcolor{road}{$\blacksquare$}road~
				}
    \\
    
			\multicolumn{6}{c}{
				\scriptsize
                    \textcolor{parking}{$\blacksquare$}parking~
				\textcolor{sidewalk}{$\blacksquare$}sidewalk~
				\textcolor{other-ground}{$\blacksquare$}other ground~
				\textcolor{building}{$\blacksquare$}building~
				\textcolor{fence}{$\blacksquare$}fence~
				\textcolor{vegetation}{$\blacksquare$}vegetation~
				\textcolor{trunk}{$\blacksquare$}trunk~
				\textcolor{terrain}{$\blacksquare$}terrain~
				\textcolor{pole}{$\blacksquare$}pole~
				\textcolor{traffic-sign}{$\blacksquare$}traffic sign			
			}
		\end{tabular}	
 	\end{center}
        \vspace{-0pt}
	\caption{ \textbf{Qualitative results} on the SemanticKITTI validation set. The overlay shadow areas at the bottom of semantic predictions denote unseen scenery out of the camera's field of view (FOV). } 
	\label{fig_q}
 \vspace{-0pt}
\end{figure*}

\begin{table}[!ht]\centering
\small
\renewcommand\tabcolsep{5.5pt}
\begin{tabular}{ l |ccc}
\toprule
 \textbf{Methods}& \textbf{Resolution}& \textbf{mAP} $\uparrow$  & \textbf{NDS} $\uparrow$   \\  \midrule

BEVDet-Base (2021) & 1600$\times$ 640 & 0.397 & 0.477 \\

BEVDet4D-Base (2021) & 1600$\times$ 640 & 0.426 & 0.552 \\

PETR-R101 (2022) & 1408$\times$ 512 & 0.357 & 0.421\\

BEVDepth-R101 (2023) & 512$\times$ 1408 & 0.412 & 0.535 \\

\rowcolor{gray!10} BRGScene (ours) & 1600$\times$ 640  &\textbf{0.451} &\textbf{0.563} \\  
  \bottomrule
\end{tabular}
\vspace{-0pt}
\caption{ \textbf{Quantitative results} of BEV Detection on the nuScenes validation set. We conduct preliminary experiments by employing the detection head. }
\vspace{-0pt}
\label{tabcom2}
\end{table}

\begin{figure}[!ht]
   \vspace{-0pt}
	\begin{center}
        \vspace{-0pt}
        \includegraphics[width=1\linewidth]{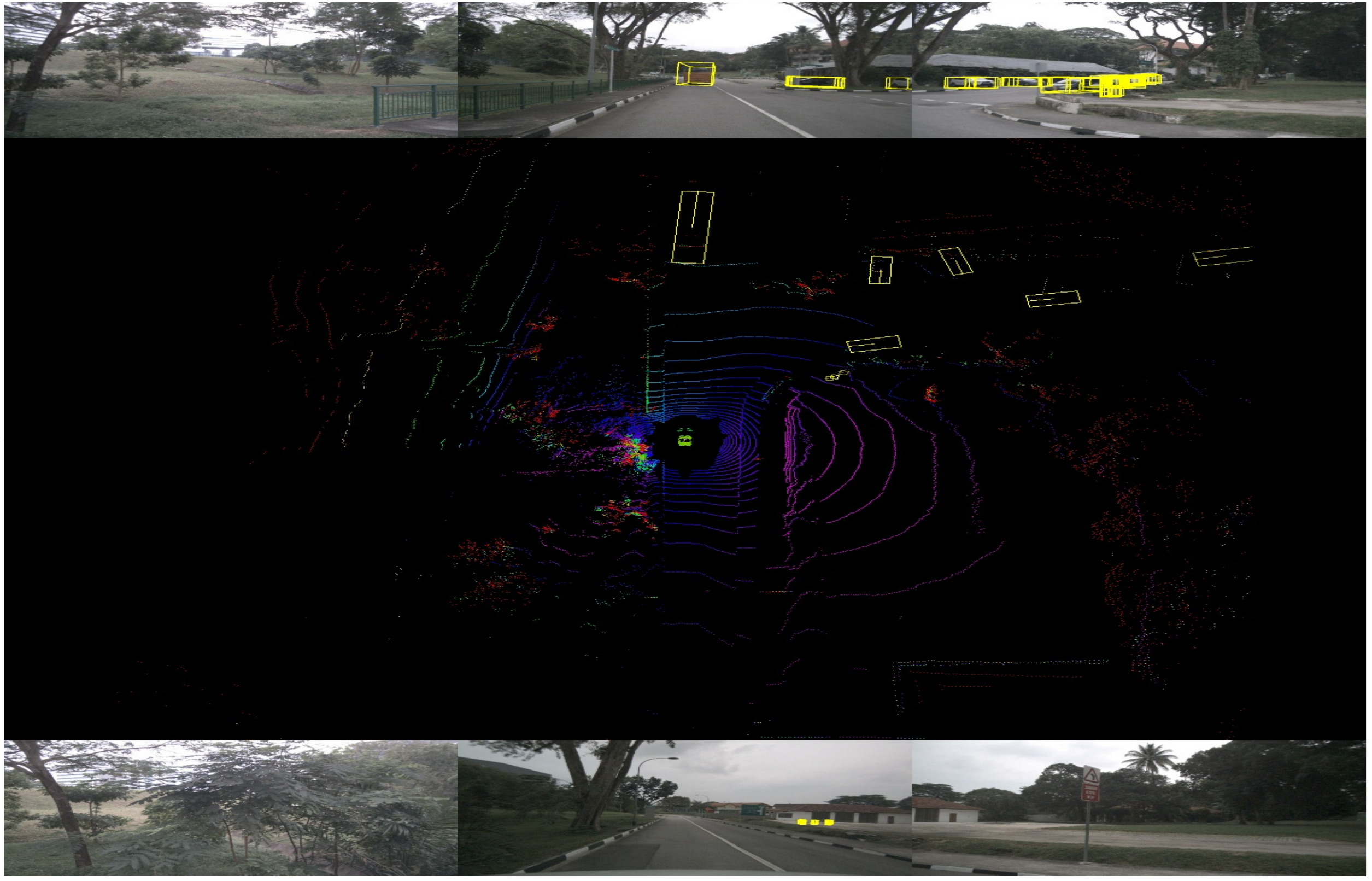}
        \end{center}
        \vspace{-0pt}
	\caption{\textbf{Visualization results} of BEV detection on the nuScenes validation set. } 
	\label{fgvisdet2}
 \vspace{-0pt}
\end{figure}

\subsection{Performance}
\noindent \textbf{Quantitative Comparison.}
Table \ref{tabq1} reports the performance of our BRGScene and other baselines on the SemanticKITTI test set. 
We compare our method with the best models~\cite{li2023voxformer,zhang2023occformer,huang2023tri,wei2023surroundocc,cao2022monoscene} for semantic scene completion. 
BRGScene surpasses temporal stereo-based (Stereo-T) VoxFormer-T~\cite{li2023voxformer} in terms of mIoU.
Additionally, our approach outperforms stereo-based VoxFormer-S by a large margin in terms of geometric completion (42.95$\rightarrow$43.34) and semantic segmentation (12.20$\rightarrow$15.36).
It's worth noting that our method demonstrates significant superiority in the prediction of small moving objects compared to VoxFormer-S, including bicycle (1.00$\rightarrow$3.40), motorcycle (0.70$\rightarrow$2.40), pole (5.10$\rightarrow$7.00), etc. We ascribe such improvements to the ensemble of dual volume, which is critical for 3D geometry awareness. 
Besides, although VoxFormer-T employs up to 4 temporal stereo image pairs as inputs, our method has a significant advantage in mIoU, with a 14.5\% relative improvement in terms of semantics.\looseness=-1

\noindent \textbf{Evaluation of Stereo Variants.}
To ensure fair comparisons, we also implement stereo variants of the baselines as shown in Table \ref{tabq3}. For MonoScene${}^{*}$~\cite{cao2022monoscene} and TPVFormer${}^{*}$~\cite{huang2023tri}, we employ left and right images to generate stereo-based predictions. 
Other LiDAR-based baselines including LMSCNet~\cite{roldao2020lmscnet} and SSCNet~\cite{song2017semantic} require 3D geometric inputs, so we adapt them with pseudo-3D inputs leveraging the same GwcNet~\cite{guo2019group} used in our framework.  
Our proposed BRGScene efficiently outperforms all the other methods. For more details on model complexity, please refer to the Supplementary Material.\looseness=-1

\noindent \textbf{Qualitative Comparison. }
 As shown in Figure~\ref{fig_q}, we compare the visualization results of BRGScene and VoxFormer on the SemanticKITTI validation set. 
 Due to the complexity of the real-world scenes and the sparsity of the labels, it is challenging to reconstruct the scenes accurately and completely. 
Compared to VoxFormer-T and VoxFormer-S, our method evidently captures better geometric representations for more complete and precise scene reconstruction (e.g. crossroads in rows 1,2) and generates more proper hallucinations outside FOV (e.g. shadow regions in rows 2,3).

\noindent \textbf{BEV Detection Evaluation.} We further conduct preliminary experimental results for BEV 3D detection on nuScenes validation set. Specifically, we adopt \textit{BEVDet}~\cite{huang2021bevdet} as the baseline setting, and replace the \textit{BEVDet} model with our proposed \textit{BRGScene} while maintaining the detection head. 
Note that we adopt temporal inputs from current and previous images to construct the temporal volume, which replaces the original stereo volume.
The preliminary results are in Table~\ref{tabcom2} and Figure~\ref{fgvisdet2}, which show that our proposed method can also be applied to a more wide range of downstream tasks.

\begin{table}[!ht]\centering
\small
\begin{tabular}{cc|cc|cc}\toprule
\multicolumn{2}{c|}{\textbf{Dual Volume}} & \multicolumn{2}{c|}{\textbf{MIE}} &\multirow{2}{*}{\textbf{IoU (\%)} $\uparrow$ } &\multirow{2}{*}{\textbf{mIoU (\%)} $\uparrow$} \\
 Stereo & BEV & BRI & DVE   \\ \midrule
  &        & &  &     33.87    &   9.92   \\
\checkmark& & &  &   37.38          &   11.49   \\
          &\checkmark & & & 36.34     &   11.06  \\ \midrule
\checkmark &\checkmark & & & 39.68 & 12.53  \\
\checkmark &\checkmark & \checkmark& & 43.04 &14.64   \\
\checkmark &\checkmark &           &\checkmark &42.92 & 14.59  \\
\rowcolor{gray!10} \checkmark &\checkmark & \checkmark &\checkmark & \textbf{43.85} & \textbf{15.43}  \\
\bottomrule
\end{tabular}
\vspace{-0pt}
\caption{ \textbf{Ablation study for architectural components.} }
\vspace{-0pt}
\label{tab_ar}
\end{table}

\subsection{Ablation Study}
We ablate our BRGScene on the SemanticKITTI validation set for semantic scene completion. 
The ablation study for the architectural components is shown in Table~\ref{tab_ar}, which includes the Dual Volume and the Mutual Interactive Ensemble block. The baseline in the first row of the table is built by removing the Dual Volume and the MIE block.

\noindent \textbf{Effect of the Dual Volume.}
For the ablation study on the dual volume, we build the framework with an individual volume to verify the effect.
We find that after adding the stereo volume, the geometric perception ability of the framework improves obviously (+3.51 IoU), while the prediction performance of semantic scene segmentation is enhanced as well (+1.57 mIoU). 
The introduction of BEV volume also has an obvious impact on the semantic and the geometric aspects
, boosting IoU by 2.47 and mIoU by 1.14, respectively.\looseness=-1

\noindent \textbf{Effect of the Mutual Interactive Ensemble.}
We further conduct architectural ablation to evaluate the impact of the Mutual Interactive Ensemble (MIE) block as shown in Table~\ref{tab_ar}. The BRI module can significantly improve the geometric and semantic estimations (+3.36 IoU, +2.11 mIoU) with efficient mutual interaction. Furthermore, we evaluate the DVE module by replacing the alternative naive concatenation, which leads to significant improvements in performance (+3.24 IoU, +2.06 mIoU).
The aforementioned results validate that our complementary mutual interaction has a significant performance improvement compared to naive aggregation.

\section{Conclusion}
In this work, we propose BRGScene, a 3D Semantic Scene Completion framework that leverages both stereo and BEV representations to produce reliable 3D scene understanding results. 
To bridge the representation gap between the stereo volume and BEV volume for fine-grained 3D perception, a {Mutual Interactive Ensemble} block is proposed to incorporate complementary merits of the two dense volumes.
Our BRGScene outperforms existing camera-based state-of-the-arts on the challenging SemanticKITTI dataset.
We hope BRGScene could inspire further research in camera-based SSC and its applications in 3D scene understanding.

\section{Acknowledgements}

{This paper is supported in part by NSFC under Grant 62302246 and ZJNSFC under Grant LQ23F010008.}

\appendix
\twocolumn[{
\centering
 \vspace{20pt}
\section*{\Large \centering Supplementary Material for BRGScene}
 \vspace{30pt}
 }]

\begin{figure*}[!ht]
	
	\begin{center}
		\includegraphics[width=0.7\linewidth]{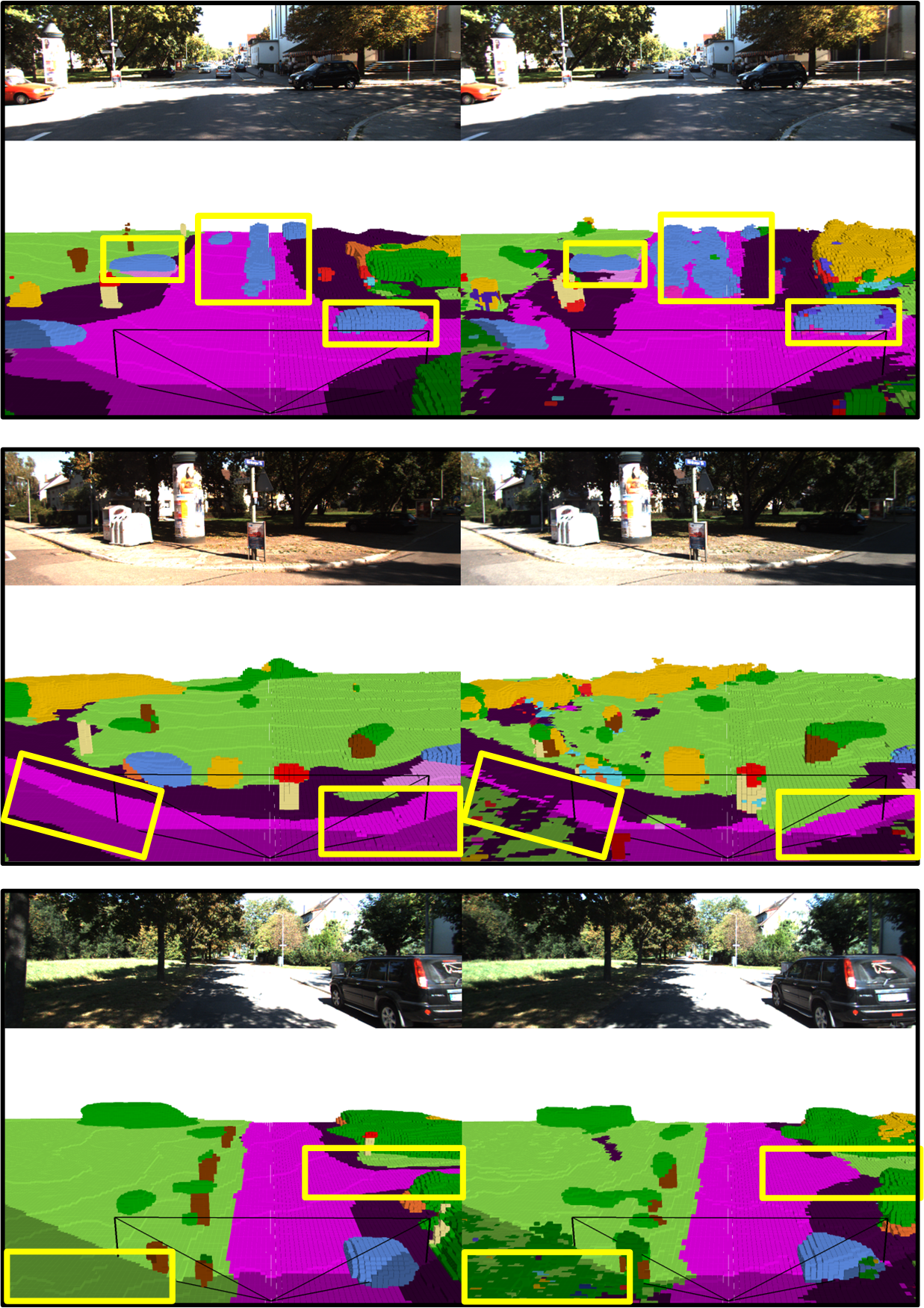}   
        \vspace{0pt}
		\begin{tabular}{cccccc}
			\multicolumn{6}{c}{
				\scriptsize
				\textcolor{bicycle}{$\blacksquare$}bicycle~
				\textcolor{car}{$\blacksquare$}car~
				\textcolor{motorcycle}{$\blacksquare$}motorcycle~
				\textcolor{truck}{$\blacksquare$}truck~
				\textcolor{other-vehicle}{$\blacksquare$}other vehicle~
				\textcolor{person}{$\blacksquare$}person~
				\textcolor{bicyclist}{$\blacksquare$}bicyclist~
				\textcolor{motorcyclist}{$\blacksquare$}motorcyclist~
				\textcolor{road}{$\blacksquare$}road~
				}
    \\
    
			\multicolumn{6}{c}{
				\scriptsize
                    \textcolor{parking}{$\blacksquare$}parking~
				\textcolor{sidewalk}{$\blacksquare$}sidewalk~
				\textcolor{other-ground}{$\blacksquare$}other ground~
				\textcolor{building}{$\blacksquare$}building~
				\textcolor{fence}{$\blacksquare$}fence~
				\textcolor{vegetation}{$\blacksquare$}vegetation~
				\textcolor{trunk}{$\blacksquare$}trunk~
				\textcolor{terrain}{$\blacksquare$}terrain~
				\textcolor{pole}{$\blacksquare$}pole~
				\textcolor{traffic-sign}{$\blacksquare$}traffic sign			
			}
		\end{tabular}	
 	\end{center}
        \vspace{0pt}
	\caption{ More visualization results on the SemanticKITTI validation set. The overlay shadow areas at the bottom of semantic predictions denote unseen scenery out of the camera's field of view (FOV). From top to below demonstrate 3 cases. For each one, the first row illustrates stereo input images while the second row depicts comparison results of our BRGScene and Voxformer-S. 
 } 
	\label{fig_video}
\end{figure*}

\begin{figure*}[!ht]
\vspace{0pt}
\hsize=\textwidth %
\centering
\includegraphics[width=0.9\textwidth]{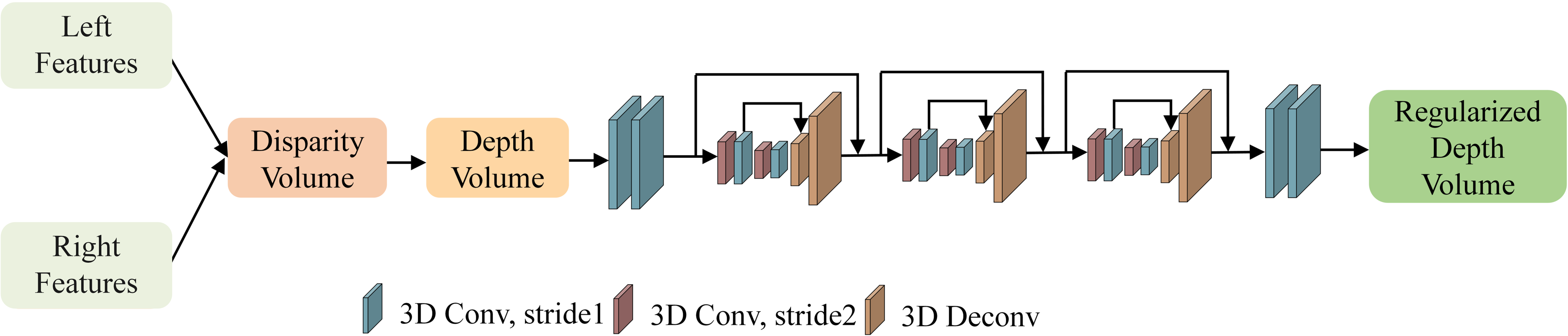}
\caption{Implementation details of the stereo representation construction. The left and right features are encoded with camera parameters before forming the disparity volume.
}
\label{figstereo}
 \vspace{0pt}
\end{figure*}

\section{More Visualization Results}
Below we provide qualitative comparisons on the SemanticKITTI validation set and show visualization examples in Figure~\ref{fig_video}.
Compared to VoxFormer-S\cite{li2023voxformer}, our {BRGScene} generates more precise predictions (e.g. cars in example 1; crossroads in example 2 and example 3) and more complete scenes with proper hallucinations (e.g. shadow regions in example 2 and example 3).

Specifically, in example 1, compared to VoxFormer-S, our method predicts the more accurate shapes for cars.
In example 2, compared to VoxFormer-S, our method successfully hallucinates a more complete and reasonable scene outside the camera FOV. 
In example 3, our BRGScene precisely captures the semantics and geometry of the crossroad and the terrain while VoxFormer-S faces challenges in correct perception.\looseness=-1

\section{Implementation Details on Dual Volume Construction}

In this section, we describe implementation details for our dual volume construction. The \emph{Stereo Geometric Volume Constructor} targets to match 2D extracted features and outputs a regularized depth volume. The \emph{BEV Latent Volume Constructor} constructs a 3D latent volume coupled with context features from 2D images.

\noindent \textbf{More Details on Stereo Representation Construction. }
As mentioned in Sec. 3.2 of the main paper, we construct stereo geometric volume from the left and right image features. 
As shown in Figure~\ref{figstereo}, the left and right image features are correlated to form the group-wise disparity volume, which is then converted into the depth volume. 
To regularize the depth volume, we adopt 3D CNNs with hourglass (encoder-decoder) structures to process it~\cite{guo2019group,cfnet}.
We first feed the depth volume to two 3$\times$3$\times$3 convolutions with stride 1. 
Next, we employ three stacked 3D hourglasses to further aggregate the volumetric information. The structure details of the 3D hourglass are shown in Table~\ref{tab:1}, where two residual connections are adopted to aggregate context from different semantic levels. In each 3D hourglass, the input volume is first downsampled and then upsampled with de-convolution, thus aggregating information along the spatial and depth dimensions. Finally, two 3$\times$3$\times$3 convolutions are used to generate the regularized depth volume. \looseness=-1

\begin{table}[!ht]\small
\begin{center}
\begin{tabular}{c|c|c|c}
\hline
Name          & Layer Properties & Dim & Input              \\ \hline
Conv1   & $3\times3\times3$, 2, 1  & $32/64$  & Input Volume    \\ \hline
Conv2   & $3\times3\times3$, 1, 1  & $64/64$  & Conv1        \\ \hline
Conv3   & $3\times3\times3$, 2, 1  & $64/128$  & Conv2      \\ \hline
Conv4   & $3\times3\times3$, 1, 1  & $128/128$  & Conv3        \\ \hline
Deconv5   & $3\times3\times3$, 2, 1  & $128/64$  & Conv4       \\ \hline
Conv6   & $1\times1\times1$, 1, 0  & $64/64$  & Conv2       \\ \hline
Shortcut  & Addition \& ReLU  & $64/64$  &\begin{tabular}[c]{@{}c@{}}Deconv5, \\ Conv6 \end{tabular}   \\ \hline

Deconv7   & $3\times3\times3$, 2, 1  & $64/32$  & Shortcut       \\ \hline
Conv8  & $1\times1\times1$, 1, 0  & $32/32$  & Input Volume       \\ \hline
Output Volume  & Addition \& ReLU  & $32/32$  & \begin{tabular}[c]{@{}c@{}}Deconv7,\\ Conv8 \end{tabular}  \\ \hline
\end{tabular}
\caption{Structure details of the 3D hourglass. The $"$Layer Properties$"$ indicates kernel size, stride and padding, respectively.\looseness=-1 }
\label{tab:1}
\end{center}
\end{table}

\begin{figure}[!ht]
\vspace{0pt}
\begin{center}		\includegraphics[width=8.5cm]{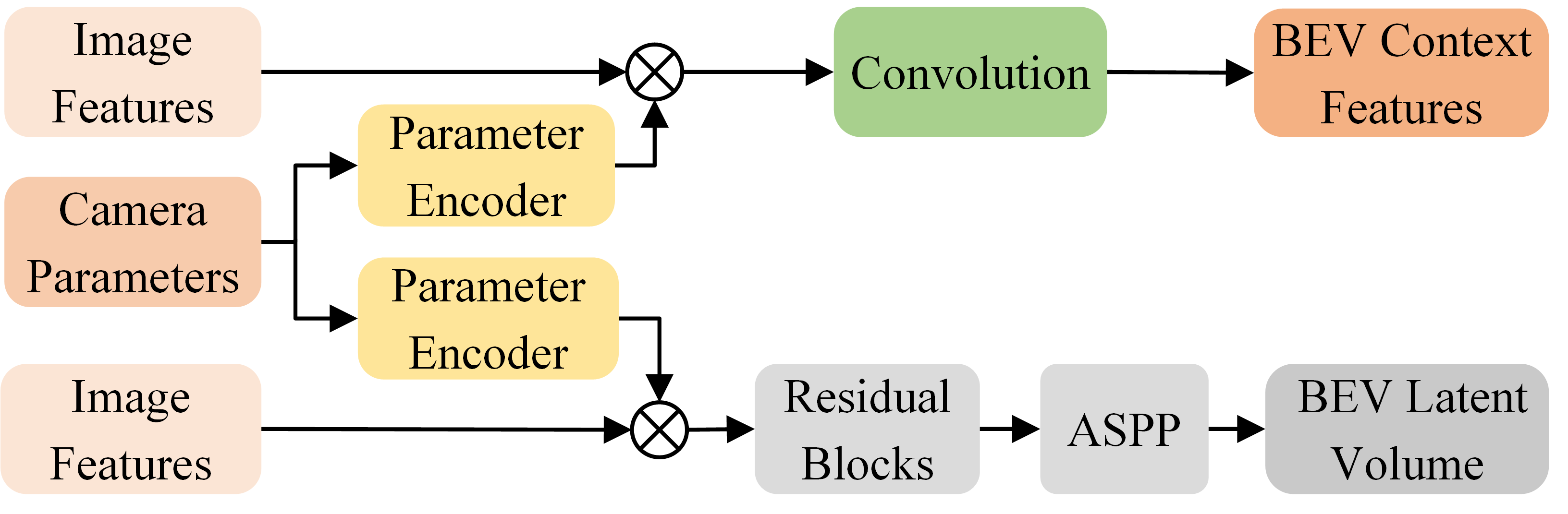}  
 	\end{center}
	\caption{ Implementation details of the BEV representation construction. We employ two branches in parallel to generate the BEV context features and BEV latent volume.} 
	\label{figbev}
 \vspace{0pt}
\end{figure}

\noindent \textbf{More Details on BEV Representation Construction.}
As mentioned in Sec. 3.2 of the main paper, we construct BEV context features and BEV latent volume with the BEV constructor~\cite{philion2020lift,li2023bevdepth}. In our implementation, we adopt left image features and camera parameters as inputs.

As shown in Figure~\ref{figbev}, two branches are employed in parallel, where the upper branch is dedicated to generating contextual features while the lower branch is dedicated to generating latent volume.  

In the parameter encoder, we concatenate camera intrinsic and extrinsic as camera parameters $P_{i}$.
For the purpose of being aware of distinct camera inputs, 
we feed the camera parameters $P_{i}$ to both branches for camera awareness similar to~\cite{li2023bevdepth}.
Formally, 

\begin{equation} \label{eq1}
P_{e}= \sigma \left ( Conv  (Reshape (FC(P_{i}))) \right )   
\end{equation}

 where $Conv$ and $FC$ are convolutions and fully-connected layers, whereas $\sigma$ and $Reshape$ represent sigmoid function and reshape operation, respectively. 
The encoded camera parameters $P_{e}$ are multiplied with the extracted features to generate the camera-aware features.

In the upper branch, a 3$\times$3 convolution is adopted to generate the context features. 
In the lower branch, we adopt two stacked residual blocks and an ASPP~\cite{chen2017rethinking,chen2018encoder} module to refine the depth distribution of the latent volume. The ASPP module aims to expand the receptive field in the depth perception process.

\begin{table}[!ht]
\vspace{-0pt}
\begin{center}
\small
\renewcommand\tabcolsep{12.0pt}
\resizebox{1.0\linewidth}{!}{
\begin{tabular}{l|cc}
\toprule
 \textbf{Methods}& \textbf{Input}  & \textbf{mIoU (\%)} $\uparrow$    \\  \midrule

MonoScene & Mono   & 11.50             \\
VoxFormer-S   &  Stereo &  12.35    \\ 
VoxFormer-T  &  Stereo-T   & 13.35   \\ 
\rowcolor{gray!10} BRGScene (ours) & Stereo   & {15.43}  \\  \midrule

SSCNet  &  LiDAR   & 16.35 \\
LMSCNet  &  LiDAR  & {17.19} \\
JS3CNet  & LiDAR   & 22.67 \\
 
 \bottomrule
\end{tabular}
}
\vspace{-0pt}
\caption{{Comparison with LiDAR-based methods} on the SemanticKITTI validation set. }
\vspace{-0pt}
\label{tab_lidar}
\end{center}
\end{table}

\section{Comparison with LiDAR-based methods.} 

We report the comparison with LiDAR-based methods in Table~\ref{tab_lidar}. 
The table illustrates that the performance gaps between our proposed method and recent LiDAR-based methods are relatively small (e.g., 0.92 mIoU drop compared to SSCNet).
These results demonstrate that our proposed stereo-based method can be promoted as a cheap alternative to LiDAR with competitive performance.

\section{More Evaluation with Current Structure}

We compare other formations of BEV volume in Tab.~\ref{tabbev}. Although stronger backbones (e.g. BEVFormer and BEVDepth) show slight improvement over LSS, the boosted performance is marginal compared with our proposed complementary volume exploitation structure.

\begin{table}[!ht]
\footnotesize
\centering
 
\begin{tabular}{l|ccc|c}
\toprule  
\textbf{Volume } & \rotatebox{0}{\scalebox{0.8}{LSS}}  & \rotatebox{0}{\scalebox{0.8}{BEVFormer}} & \rotatebox{0}{\scalebox{0.8}{BEVDepth}} & \rotatebox{0}{\scalebox{0.8}{LSS+Stereo}}  
\\ \midrule
 
\textbf{IoU(\%)} $\uparrow$ & 38.97   & 39.33 & 39.42 & \textbf{43.85} \\

\textbf{mIoU(\%)} $\uparrow$  & 13.18   &  13.44 & 13.51 & \textbf{15.43} \\ 

 \bottomrule
\end{tabular}
\caption{{The effect of volume representations with different BEV baselines.}} 
\centering
\label{tabbev}
\end{table}

\vspace{0pt}
\bibliographystyle{named}
\bibliography{ijcai24}

\end{document}